\documentclass[letterpaper]{article} 
\usepackage{aaai2026}  
\makeatletter
\gdef\copyright@on{}
\makeatother
\usepackage{times}  
\usepackage{helvet}  
\usepackage{courier}  
\usepackage[hyphens]{url}  
\usepackage{graphicx} 
\usepackage{natbib}  
\usepackage{caption} 
\usepackage{booktabs}
\usepackage{multirow}
\usepackage{amssymb}
\usepackage{amsmath}

\usepackage{algorithm}
\usepackage{algorithmic}

\usepackage{newfloat}
\usepackage{listings}
\DeclareCaptionStyle{ruled}{labelfont=normalfont,labelsep=colon,strut=off} 
\floatstyle{ruled}
\newfloat{listing}{tb}{lst}{}
\floatname{listing}{Listing}
\title{Spatial-Frequency Aware for Object Detection in RAW Image}
\author {
    Zhuohua Ye\textsuperscript{\rm 1},
    Liming Zhang\textsuperscript{\rm 1},
    Hongru Han\textsuperscript{\rm 1}
}
\affiliations {
    \textsuperscript{\rm 1}University of Macau\\
    yc37977@um.edu.mo, 
    lmzhang@um.edu.mo, 
    yc37462@um.edu.mo
}

\begin{document}

\maketitle

\begin{abstract}
Direct RAW-based object detection offers great promise by utilizing RAW data (unprocessed sensor data), but faces inherent challenges due to its wide dynamic range and linear response, which tends to suppress crucial object details. In particular, existing enhancement methods are almost all performed in the spatial domain, making it difficult to effectively recover these suppressed details from the skewed pixel distribution of RAW images. To address this limitation, we turn to the frequency domain, where features, such as object contours and textures, can be naturally separated based on frequency. In this paper, we propose Space-Frequency Aware RAW Image Object Detection Enhancer (SFAE), a novel framework that synergizes spatial and frequency representations. Our contribution is threefold. The first lies in the ``spatialization" of frequency bands. Different from the traditional paradigm of directly manipulating abstract spectra in deep networks, our method inversely transforms individual frequency bands back into tangible spatial maps, thus preserving direct physical intuition. Then the cross-domain fusion attention module is developed to enable deep multimodal interactions between these maps and the original spatial features. Finally, the framework performs adaptive nonlinear adjustments by predicting and applying different gamma parameters for the two domains.
\end{abstract}


\section{Introduction}
\begin{figure}[t]
  \centering
  \includegraphics[width=0.49\textwidth]{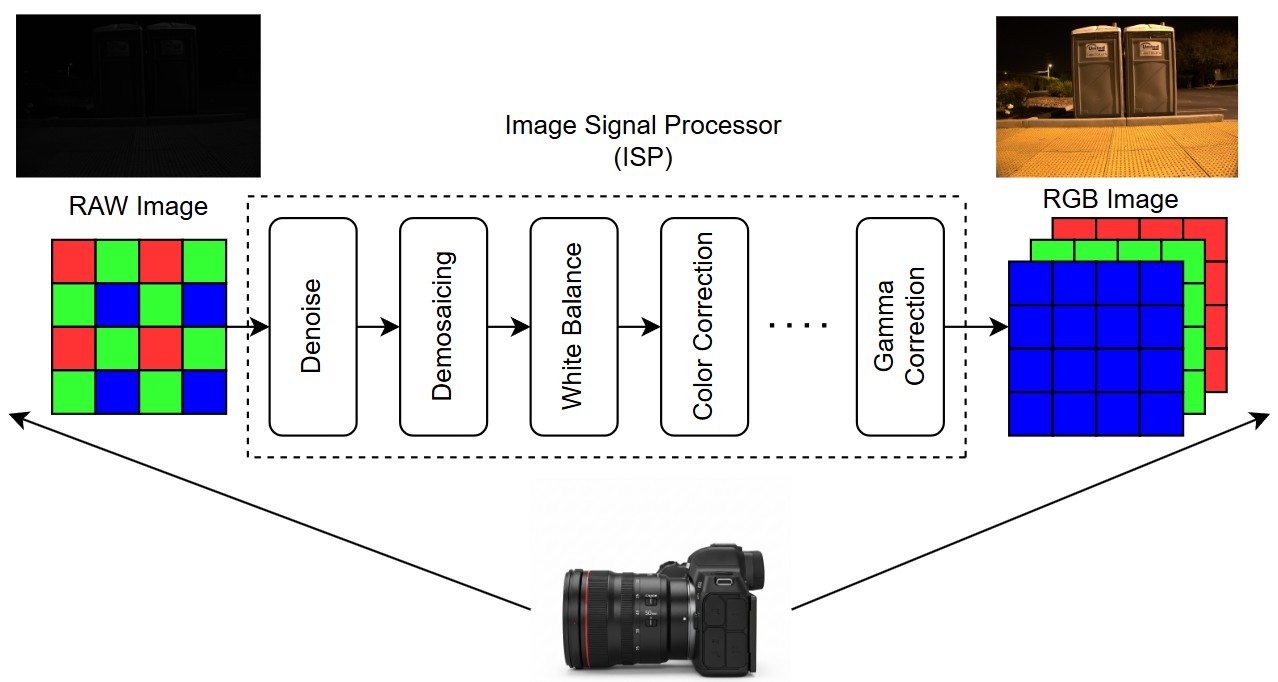}
  \caption{Illustration of the Image Signal Processor (ISP) inside a camera.}
  \label{fig:ISP}
\end{figure}

RAW images, as shown in Figure \ref{fig:ISP}, are the direct output of camera sensors. Specifically, they are captured as a single-channel mosaic of light intensities, where each photosite is filtered to receive red, green, or blue photon, commonly arranged in a Bayer pattern (e.g., RGGB). Thus, while technically single-channel, a RAW image contains the unprocessed signals for all three primary colors. This primitive format retains the complete information of the scene \citep{Chen_2018_CVPR,zamir2020cycleisp} and serves as the underlying source for standard RGB (sRGB). This also makes it a theoretically optimal medium for high-level vision tasks such as object detection. However, the RAW data faces great challenges in practice, because its high dynamic range and linear characteristics lead to skewed pixel distribution. Most of the pixel information in RAW data is concentrated in the low or high intensity range \citep{raworcooked,li2024efficient}, which causes the texture details of the object to be suppressed in the images. 

The research for processing RAW data in the literature is mainly divided into two streams. One stream only focuses on the spatial domain and aims to replace the traditional image signal processor (ISP). Research in this direction has evolved from early U-Net-based models \citep{Chen_2018_CVPR} to more complex architectures \citep{dai2020awnet} that employ multi-scale processing \citep{ignatov2020replacing} and wavelet transforms \citep{lamba2021restoring}, and subsequently focused on lightweight designs \citep{chen2022lw, ignatov2022mobilegpu, ignatov2021mobilenpu} for real-time performance. Another mainstream research explores hybrid methods that combine deep networks with interpretable ISP modules, such as learning to ``unprocess" sRGB images through ISP and denoising based on learning to simulate RAW data \citep{brooks2019unprocessing}, or using the network to predict the parameters of the traditional ISP module \citep{conde2022modelbased}, or mapping the image to a device-independent space for processing through a network-based ISP and then re-rendering it to sRGB \citep{CIEXYZNet}. 

While the above streams of research all start from RAW data, they serve a very different goal. The RAW-to-sRGB conversion pipeline is tailored for the human eye, employing complex operations, as in Figure \ref{fig:ISP}, to produce perceptually pleasing images. However, what is optimal for human perception is not necessarily optimal for machine vision. This contradiction is particularly evident with sensor noise. Although noise removal is critical for human vision, the foundational work of Buckler et al. \citep{buckler2017reconfiguring} suggests that such RAW noise may have little impact on high-level vision tasks. This implies that human-centric processing, such as denoising, may not be necessary for machine learning models, or even degrade the information quality (e.g., through over-smoothing). This motivates our work, which is dedicated to finding effective ways specifically for machine vision-oriented tasks, rather than simply improving the visual quality of sRGB images.

Since textures and fine details of objects naturally correspond to mid- and high-frequency components of an image \citep{oppenheim1975digital}, a method that can selectively act on these frequency bands is needed to recover the hidden information. This is where the frequency domain perspective becomes invaluable. Unfortunately, almost all existing methods for enhancing RAW data for object detection operate only in the spatial domain. These methods process all pixel information uniformly, making it difficult for them to separate the key frequency components for object detection from tilted RAW data, which often leads to unstable training and inefficient feature extraction. 
\begin{figure}[t]
  \centering
  \includegraphics[width=0.45\textwidth]{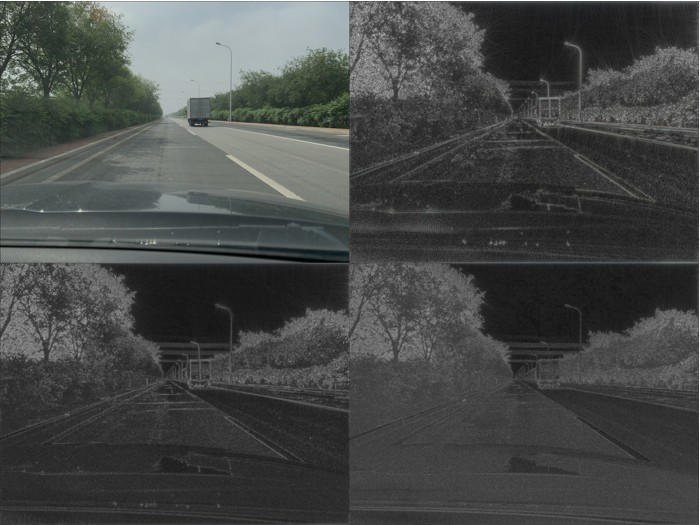}
  \caption{The visualization of spatialized frequency band maps. From top-left to bottom-right, the components are arranged from low frequency to high frequency.}
  \label{fig:Spatialized-frequency-bands}
\end{figure}

In the field of conventional computer vision, frequency domain information has proven to be powerful for various visual tasks. Techniques such as Fast Fourier Convolution (FFC) have been used to efficiently obtain global receptive fields and have achieved state-of-the-art (SOTA) performance in applications such as image inpainting \citep{suvorov2022resolution}. Although frequency information has also been introduced into RAW data processing, such as using Fourier domain to separate style and structure to enhance RAW to sRGB conversion \citep{fftenhancing}, these methods usually use the Fourier spectrum directly, resulting in a domain gap with spatial features. The motivation of our method is to bridge this gap by introducing spatialized frequency band maps, giving tangible spatial semantics to the frequency components. Furthermore, we focus on developing targeted and learnable modules that build on previous foundational research, in particular, Buckler et al. \citep{buckler2017reconfiguring}, who questioned the necessity of a complete ISP pipeline for machine vision, arguing that only certain components, especially nonlinear transformations like gamma correction, are necessary. Subsequent research on object detection based on RAW format further confirmed this view and further highlighted the key role of such adjustments in improving task performance \citep{raworcooked, CVPR2023_ROD}.

In this paper, we propose a novel RAW image enhancement framework: \textbf{Spatial-Frequency Aware Enhancer (SFAE)}. Our method introduces a parallel two-stream architecture to simultaneously process the raw spatial image and a novel representation we call spatialized frequency band maps. These maps are generated by converting the band-limited spectrum back to the spatial domain, giving the abstract frequency components a concrete spatial semantic, as shown in Figure~\ref{fig:Spatialized-frequency-bands}. The core of our model is a cross-domain attention fusion module. Inspired by multimodal studies that fuses text domain and images domain \citep{crossdomain}, we establish a multimodal dialogue between spatial domain and frequency domain. This enables the global spatial context to guide the enhancement of key frequency details and vice versa, thereby achieving deep collaboration and complementarity of cross-domain information to truly unleash the potential of RAW data. In addition, we extend the whole-image nonlinear gamma correction \citep{buckler2017reconfiguring,raworcooked,li2024efficient} into fine-grained adaptive gamma correction, which predicts and applies independent gamma parameters for the spatial domain image and each individual spatialized frequency band map, thus achieving highly targeted enhancement.

\noindent Our contributions are summarized as follows.
\begin{itemize}
    \item We introduce Spatialized Frequency Band Maps, a novel representation that bridges the gap between abstract frequency information and intuitive spatial features, enabling the targeted recovery of details suppressed in RAW data.

    \item We introduce the concept of multimodal learning into RAW based object detection via cross-domain attention fusion mechanism, leveraging complementary information from different domains (e.g., frequency and spatial) to enhance detection performance.

    \item We propose a novel dual-domain adaptive enhancement strategy. Our framework learns to apply distinct gamma corrections not only to the spatial image but also, uniquely, to each individual spatialized frequency band map, providing highly fine-grained, content-aware control for optimal enhancement toward detection task.
\end{itemize}

\section{Related Work}
\subsection{RAW data processing for high-level visual tasks}
The early development of computer vision was deeply tied to sRGB. Taking advantage of large-scale sRGB datasets such as ImageNet \citep{deng2009imagenet} and a series of deep models built on them \citep{ren2015faster}, computer vision has made great progress. Early RAW research focused on generating visually pleasing sRGB images for human perception. However, its performance in challenging scenarios is limited due to information loss during ISP processing. Subsequent research began to customize ISPs for specific tasks. These hybrid approaches include using neural networks to predict the parameters of traditional ISP modules, such as \citep{morawski2022genisp,raw_adapter,IA-ISP}. Despite the innovations of these methods, they still suffer from the problem of potential error accumulation during serial processing. This has prompted a shift to directly using richer RAW data for advanced vision tasks.

\subsection{Simplification of ISP for high-level vision tasks}
The role of traditional ISP in machine vision was first questioned by Buckler et al. \citep{buckler2017reconfiguring}. The core finding is that demosaic and gamma correction are the two critical steps in the image classification task. Ljungbergh et al. \citep{raworcooked}, who propose learnable Yeo-Johnson transformation, deepen this exploration from classification tasks to more complex detection tasks. They experimented in PASCALRAW dataset \citep{pascalraw}. The experimental results strongly support Buckler et al.'s conclusion. In the work \citep{CVPR2023_ROD}, they combine the gamma correction with their proposed pixel-level nonlinear transformation method, which also reinforces the significance of nonlinear transformation in RAW based object detection task. Lu et al. \citep{lu2023enhancing} even replace the whole ISP with their proposed non-learning based method named Log-Gradient in the task of RAW based image classification, which surprisingly have a superior performance.

\subsection{Frequency domain in computer vision task}
Fourier transform can decompose an image into components of different frequencies, which physically correspond to different attributes of the image \citep{oppenheim1975digital}. In deep learning, the use of frequency domain has become an active research direction. In the field of image restoration, Li et al. \citep{Li2023ICLR} introduce Fourier layer into the network to better model global illumination. Zhou et al. \citep{sun2024fourmer} designed a Transformer based frequency domain model for image restoration, demonstrating the great potential of frequency domain modeling. 

In existing methods, the results of frequency domain analysis (such as spectrograms, phase and amplitude) are often directly fed into the network, exhibiting a representational gap with spatial domain features. In this paper, we introduce the spatialized frequency band maps in spatial domain and design a spatial-frequency cross-fusion attention for object detection.

\section{Method}
\begin{figure*}[t]
  \centering
  \includegraphics[width=\textwidth]{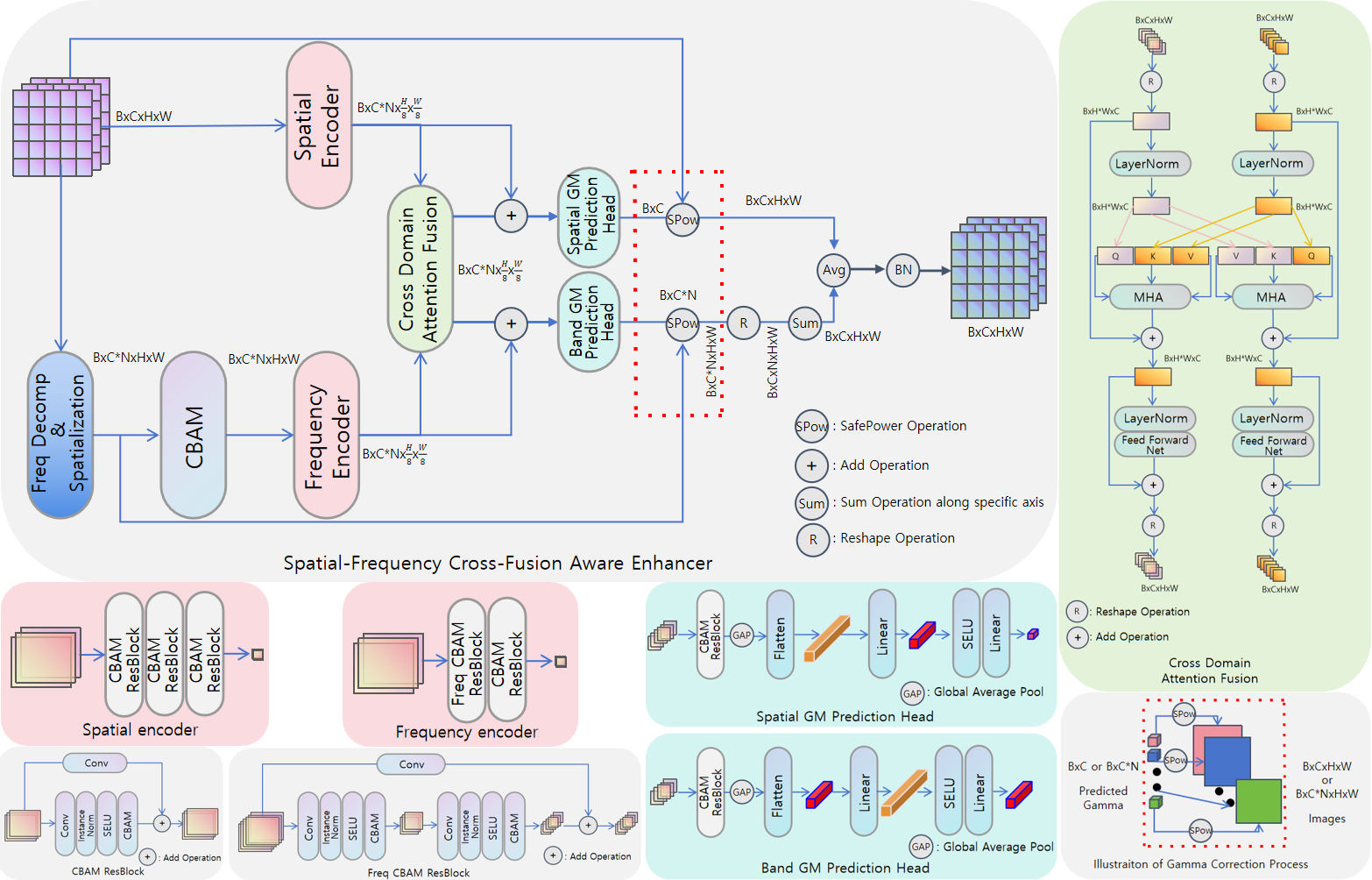}
  \caption{Overview of the proposed SF-CFAE framework.  The Spatial and Frequency Encoder, Cross Domain Attention Fusion, Band and Spatial GM Prediction Head modules are further described in detail in the sub-figures corresponding to the same colors. The dotted squares illustrate the gamma correction process.}
  \label{fig:Model-architecture}
\end{figure*}

The structure of our proposed SF-CFAE is illustrated in Figure \ref{fig:Model-architecture}, 
which consists of four parts: (1) Frequency Decomposition and Spatialization module; (2) Spatial-Frequency dual-branch Encoder; (3) Cross-Domain Attention Fusion module; and (4) Frequency band Adaptive Nonlinear Enhancement module.

\subsection{Frequency Decomposition and Spatialization}
Our methodology is rooted in a core insight: the physical contents of an image can be transformed into different frequencies in the frequency domain. To take advantage of this feature, we first need to decompose RAW images $I_{\text{RAW}} \in \mathbb{R}^{B \times C \times H \times W}$ into different bands where B is the batch size, C is the number of channels, H, W is the height and width.
First, we apply two-dimensional discrete Fourier transform (DFT) to transform the image from spatial domain to frequency domain:
\begin{equation}
    F(u, v) = \sum_{x=0}^{H-1} \sum_{y=0}^{W-1} I_{\text{RAW}}(x, y) \cdot e^{-j 2 \pi (\frac{ux}{H} + \frac{vy}{W})}.
\end{equation}
\begin{itemize}
\item{\textbf{Band division:}} Inspired by octave analysis in signal processing \citep{octave}, we divide the two-dimensional frequency space into $N$ annular frequency bands on a logarithmic scale. The boundaries for each band $i$ (for $i=1, \dots, N$), defined by normalized radii $[f_{\text{low},i}, f_{\text{high},i})$, are generated from a maximum frequency $f_{\text{max}}$ (typically 0.5). The upper bound is given by $f_{\text{high},i} = f_{\text{max}} / 2^{N-i}$, while the lower bound is the upper bound of the preceding band, $f_{\text{low},i} = f_{\text{high},i-1}$ for $i>1$, with the first band starting at zero ($f_{\text{low},1}=0$).
\item{\textbf{Frequency filtering and spatialization:}} For each frequency band $B_i$, we construct a binary mask $M_i$. The value of this mask at frequency coordinate $(u, v)$ is defined as follows: if the corresponding frequency $\rho = \sqrt{u^2 + v^2}$ lies within the interval $[f_{\text{low},i},\ f_{\text{high},i})$, then $M_i(u, v) = 1$; otherwise, $M_i(u, v) = 0$. We then apply this mask to the centralized frequency spectrum $F_{\text{shift}}$ to isolate the spectral components in the $i$-th band:
\begin{equation}
F_{\text{shift},i} = F_{\text{shift}} \odot M_i \ .
\end{equation}
where $\odot$ denotes element-wise multiplication.
\end{itemize}

In our method, we do not use frequency information directly like existing methods. Instead, we reproject each band-limited frequency component $F_{\text{shift},i}$ back into the spatial domain. This is done via the inverse Fourier transform (\text{iFFT}), which first requires \text{iFFTShift} to move the zero-frequency component from the center back to the corner of the spectrum, yielding a set of Spatialized Frequency Band Maps $I_{\text{freq},i}$:
\begin{equation}
I_{\text{freq},i} = \text{iFFT}(\text{iFFTShift}(F_{\text{shift},i})) \in \mathbb{R}^{B \times C \times H \times W}.
\end{equation}
This operation is critical because it gives the abstract frequency information a concrete spatial semantics. For example, $I_{\text{freq},\text{high}}$ highlights the edges, noise and fine textures of the images, while $I_{\text{freq},\text{low}}$ appears blurred and captures the overall structure and illumination of the image.

Finally, we obtain a set of $N$ Spatialized Frequency Band Maps, denoted as $\{I_{\text{freq},i}\}_{i=1}^N$. Each map $I_{\text{freq},i}\ \in \mathbb{R}^{B \times  C \times H \times W}$ represents a distinct spatialized frequency band map of the original image.

\subsection{Spatial-frequency dual-branch Encoder}
To simultaneously leverage both the macroscopic spatial structure and the microscopic frequency characteristics of the image, we design a dual-branch parallel architecture. 
\begin{itemize}
\item \textbf{Spatial Branch:} This branch directly processes the RAW input image $I_{\text{RAW}}$. We employ a Spatial Encoder, which consists of stacked three CBAM ResBlocks to extract hierarchical spatial features. The encoder gradually increases the receptive field through downsampling, capturing multiscale spatial information ranging from local textures to global context. Ultimately, it produces deep spatial features $z_{\text{spa}} \in \mathbb{R}^{B \times (N \cdot C) \times \frac{H}{8} \times \frac{W}{8}}$.

\item \textbf{Frequency Branch:} This branch processes the set of $N$ spatialized frequency band maps $\{I_{\text{freq},i}\}_{i=1}^N$. For practical implementation, these are concatenated along the channel dimension to form a single input tensor of shape $\mathbb{R}^{B \times (N \cdot C) \times H \times W}$. A Freqency Encoder stacked with one Freq CBAM ResBlock and one CBAM ResBlock is used to extract deep frequency features $z_{\text{freq}} \in \mathbb{R}^{B \times (N \cdot C) \times \frac{H}{8} \times \frac{W}{8}}$.
\end{itemize}
In particular, CBAM \citep{cbam} Attention Module is embedded within both CBAM ResBlock and Freq CBAM ResBlock, which are used to construct the Frequency and Spatial encoders to adaptively emphasize the most informative frequency bands and spatial regions. Furthermore, the Instance Norm \citep{in} and SELU \citep{selu} activation function are applied in series before CBAM and after Convolution.
\subsection{Cross-Domain Attention Fusion} 
Simply adding or concatenating $z_{\text{spa}}$ and $z_{\text{freq}}$ is an inefficient fusion strategy. To enable a deeper interaction between the two domains, we adopt a multimodal fusion paradigm. We treat the spatial features and our unique spatialized frequency representations as two distinct modalities, fusing them with a Cross-Domain Attention Fusion module. This module includes two parallel Multi-Head Attention (MHA) \citep{transformer} operations:
Frequency-Guided MHA for the input of spatial domain feature uses $LN(z_{\text{spa}})$ as the query (\textit{q}) and $LN(z_{\text{freq}})$ as the key (\textit{k}) and value (\textit{v}), the frequency-enhanced spatial domain features $z_{\text{spa}}'$ are computed as:
\begin{equation}
z_{\text{spa}}' = \text{MHA}(LN(z_{\text{spa}}),\ LN(z_{\text{freq}}),\ LN(z_{\text{freq}})) + z_{\text{spa .}}
\end{equation} 
where LN denotes LayerNorm \citep{layernorm}. This process allows global frequency signals (e.g. texture intensity, noise distribution) to guide the spatial feature map, enabling it to focus on regions critical for the detection task.
In contrast, the Spatial-Guided MHA for inputs of frequency domain features uses $LN(z_{\text{freq}})$ as \textit{q} and $LN(z_{\text{spa}})$ as \textit{k} and \textit{v}, the spatial-context-enhanced frequency domain features $z_{\text{freq}}'$ are computed as:
\begin{equation}
z_{\text{freq}}' = \text{MHA}(LN(z_{\text{freq}}),\ LN(z_{\text{spa}}),\ LN(z_{\text{spa}}))+z_{\text{freq .}}
\end{equation}
This allows the model to dynamically adjust the emphasis on different frequency components based on spatial layout and object contours. Both $z_{\text{freq}}'$ and $z_{\text{spa}}'$ are subsequently processed by the following network consisting of a LayerNorm and a Feed Forward Net \citep{ffn} with a skip connection as shown in Figure \ref{fig:Model-architecture}. Through this cross-domain attention fusion mechanism, the features from the two branches are no longer processed independently but rather achieve deep synergy and complementarity.
\subsection{Frequency band Adaptive Nonlinear Enhancement}
Nonlinear transformations (e.g., gamma correction) are crucial to unleash the potential of RAW data. Instead of only applying transformation to spatial image, our method also aims to achieve band-level adaptive enhancement. We design two independent prediction heads, taking the fused features $z_{\text{spa}}'$ and $z_{\text{freq}}'$ as input respectively. One predicts a channel-wise gamma value for the original spatial domain image: $\gamma_{\text{orig}} \in \mathbb{R}^{B \times C \times 1 \times 1}$. The other predicts a band-channel-wise gamma value for each spatialized frequency band map: $\gamma_{\text{freq}} \in \mathbb{R}^{B \times (N \cdot C) \times 1 \times 1}$.
\\

\noindent We then apply these adaptive gamma corrections using a numerically stable power function, denoted as $\mathrm{SafePow}(S,\ \gamma)$. For a signal $S$, the enhanced signal $S'$ is computed as:
\begin{equation}
S' =\mathrm{SafePow}(S,\ \gamma) = \text{sign}(S) \cdot (|S| + \epsilon)^{\gamma}.
\end{equation}
where $\epsilon$ is a small constant (e.g., $1\text{e}{-6}$) to prevent gradient issues when raising zero to a power, and $\gamma$ is the adaptively predicted gamma parameter from our network.

\noindent As a result, we obtain two sets of enhanced outputs:
\begin{itemize}
  \item Enhanced original image:
\begin{equation}
  I'_{\text{orig}} = \mathrm{SafePow}(I_{\text{RAW}},\ \gamma_{\text{orig}}).
\end{equation}
  \item Enhanced frequency band maps, applied to each band individually:
\begin{equation}
  I'_{\text{freq}, i} = \mathrm{SafePow}(I_{\text{freq}, i}, \gamma_{\text{freq}, i}), \quad \text{for } i=1, \dots, N.
\end{equation}
\end{itemize}
The final step is to recombine the enhanced multi-branch information into a single, optimized image. We first aggregate the set of enhanced frequency band maps $\{I'_{\text{freq}, i}\}_{i=1}^N$ by summing them to obtain a unified frequency-enhanced image $I'_{\text{freq,sum}} \in \mathbb{R}^{B \times C \times H \times W}$:
\begin{equation}
I_{\text{freq, sum}}' = \sum_{i=1}^{N} I_{\text{freq}, i\space .}'
\end{equation}

\noindent Next, this aggregated frequency-enhanced map is fused with the spatial-enhanced image $I_{\text{orig}}'$ by averaging to produce the final enhanced image:

\begin{equation}
I_{\text{enhanced}} = \mathrm{AVG}(I_{\text{orig}}',\ I_{\text{freq, sum}}').
\end{equation}

\noindent This resulting image is subsequently fed into any standard downstream object detector (e.g., YOLO \citep{redmon2016you}, Faster R-CNN \citep{ren2015faster}) for training and inference.

\section{Experiment}
\begin{table*}[t]
\centering
\small
\begin{tabular}{@{}l|ccc|ccc|ccc@{}}
\toprule
\multirow{2}{*}{\textbf{Method}} & \multicolumn{3}{c|}{\textbf{LOD}} & \multicolumn{3}{c|}{\textbf{NOD-Nikon}} & \multicolumn{3}{c}{\textbf{NOD-Sony}} \\
& \multicolumn{3}{c|}{\small{(RetinaNet R50)}} & \multicolumn{3}{c|}{\small{(Sparse R-CNN R50)}} & \multicolumn{3}{c}{\small{(Sparse R-CNN R50)}} \\
\cmidrule(l){2-10}
& mAP & mAP\textsubscript{50} & mAP\textsubscript{75} & mAP & mAP\textsubscript{50} & mAP\textsubscript{75} & mAP & mAP\textsubscript{50} & mAP\textsubscript{75} \\
\midrule 
sRGB  & 35.9 & 56.2 & 39.4 & 25.9 & 46.1 & 25.7 & 28.1 & 50.5 & 27.6 \\
RAW  & 34.0 & 54.6 & 35.3 & 14.3 & 32.4 & 11.6 & 28.3 & 49.6 & 28.0 \\
GenISP   & 34.9 & 57.9 & 33.9 & 27.7 & 49.5 & 27.0 & \underline{31.0} & 52.7 & 30.8 \\
Log-Gradient   & 30.4 & 51.3 & 31.5 & 24.2 & 43.6 & 27.3 & 25.3 & 46.7 & 23.9 \\
Yeo-Johnson   & 36.8 & 58.0 & \underline{40.1} & 26.4 & 47.3 & 25.3 & 28.0 & 49.9 & 27.9 \\
Adaptive Module   & 37.0 & 58.5 & 38.4 & 28.3 & 49.5 & 28.2 & 29.9 & 52.5 & 29.5 \\
RAW-Adapter   & 37.0 & 58.4 & 38.5 & 26.2 & 47.7 & 25.0 & 30.7 & 52.7 & \textbf{31.4} \\
IA-ISP   & \underline{37.2} & \textbf{59.9} & 38.6 & \underline{29.3} & \underline{50.8} & \underline{29.4} & 30.5 & \underline{52.8} & 31.1 \\
\midrule 
\textbf{Ours} & \textbf{39.7} & \underline{59.1} & \textbf{44.9} & \textbf{29.7} & \textbf{51.5} & \textbf{29.8} & \textbf{31.8} & \textbf{54.0} & \underline{31.3} \\
\bottomrule
\end{tabular}
\caption{Quantitative comparison with baselines and SOTA methods on the LOD and NOD datasets. We report COCO metrics: mAP (mean Average Precision across IOU from 50\% to 95\%), mAP\textsubscript{50} (mAP at 50\% IOU), and mAP\textsubscript{75} (mAP at 50\% IOU). All methods are trained and tested on the specified detector for each dataset. The best results in each column are highlighted in \textbf{bold} and the second-best are \underline{underlined}. RetinaNet and Sparse R-CNN R50 refers to RetinaNet and Sparse R-CNN with backbone ResNet-50 respectively.}
\label{tab:lod_nod_results_final_v2}
\end{table*}
To comprehensively evaluate our proposed SFAE, we conduct extensive experiments on five publicly available RAW image datasets. These datasets cover different imaging sensors, lighting conditions, object categories, and annotation characteristics, enabling us to fully test the effectiveness and generalizability of our method in various scenarios.
\subsection{Implementation Details \& Datasets}
We conduct our evaluations via MMdetection \citep{mmdetection} on datasets: LOD \citep{Hong2021Crafting}, NOD-Nikon and Sony \citep{morawski2022genisp}, and MultiRAW-Huawei and iPhone \citep{li2024efficient}. Detail of these datasets are summarized in the section Specification of Datasets in Appendix.

To demonstrate the wide applicability of our method, we train RetinaNet \citep{lin2017focal} and Sparse R-CNN \citep{sun2021sparse} with ResNet-50 \citep{He2015} as backbone on LOD and NOD datasets respectively, and also train YOLOV8-m \cite{yolov8_ultralytics} on MultiRAW dataset. The image pre-process, hyperparameter and training setup are shown in the section Experiment Setup of Appendix.

\subsection{Experiment Results \& Analysis}
We compare our method with sRGB baseline and selected SOTA RAW based object detection methods, including GenISP \citep{morawski2022genisp}, Yeo-Johnson Transform\cite{raworcooked}, RAW-Adapter \citep{raw_adapter}, Adaptive Module \citep{CVPR2023_ROD}, Log-Gradient \citep{lu2023enhancing}, and IA-ISP \citep{IA-ISP}. 

As demonstrated in Table \ref{tab:lod_nod_results_final_v2}, our proposed method, SFAE, achieves competitive results against other competitors across the three dark environment datasets. In particular, our method achieves the highest performance across all three datasets in terms of mAP, the most critical metric. Another key observation from the baseline results is the instability of directly using RAW data for object detection. Although in some scenarios, such as on the NOD-Sony dataset, the minimally processed RAW (28.3 mAP) can slightly outperform the sRGB Baseline (28.1 mAP), it can also lead to a catastrophic performance collapse. This is evident in the results of the NOD-Nikon dataset, where the RAW achieves only 14.3 mAP, a dramatic drop of over 11 points compared to its sRGB counterpart. This volatility shows that RAW data, despite containing richer information, is not inherently ``machine-friendly". 
\begin{table}[t] 
\centering
\small 
\setlength{\tabcolsep}{3pt} 
\begin{tabular}{@{}l|ccc|ccc@{}}
\toprule
\multirow{2}{*}{\textbf{Method}} & \multicolumn{3}{c|}{\textbf{MultiRAW-Huawei}} & \multicolumn{3}{c}{\textbf{MultiRAW-iPhone}} \\ 
\cmidrule(l){2-7}
& mAP & mAP\textsubscript{50} & mAP\textsubscript{75} & mAP & mAP\textsubscript{50} & mAP\textsubscript{75} \\
\midrule
sRGB & 17.3 & 30.6 & 16.7 & 29.1 & 47.5 & 29.7 \\
RAW & 15.4 & 27.3 & 14.9 & 28.8 & 46.9 & 28.8 \\
GenISP   & 18.1 & 28.4 & 17.6 & 27.9 & 45.9 & 27.7 \\
Log-Gradient   & 17.0 & 29.5 & 15.7 & 25.5 & 42.9 & 25.1 \\
Yeo-Johnson & \textbf{18.4} & 31.5 & \underline{17.8} & 29.5 & 48.0 & 30.1 \\
Adaptive Module & \underline{18.3} & \textbf{32.4} & \underline{17.8} & 29.5 & 47.7 & 30.0 \\
RAW-Adapter & --- & --- & --- & \underline{30.4} & \underline{49.8} & \textbf{31.6} \\
IA-ISP & \textbf{18.4} & 31.7 & \underline{17.8} & 30.1 & 49.2 & 30.2 \\
\midrule
\textbf{Ours} & \underline{18.3} & \underline{32.2} & \textbf{17.9} & \textbf{30.7} & \textbf{50.2} & \underline{31.2} \\
\bottomrule
\end{tabular}
\caption{Quantitative comparison on the MultiRAW datasets using YOLOv8-m. These datasets feature bright scenes with dense, small objects. The best results are in \textbf{bold}, second-best are \underline{underlined}. The dashed line (``---'') denotes NaN.}
\label{tab:multi_raw_results}
\end{table}

As demonstrated in Table \ref{tab:multi_raw_results}, while testing in a bright dataset with extreme small and dense objects, our method still achieves a performance with slight advantages in certain metrics, but consistently higher than the baseline performance of sRGB in all metrics (mAP, mAP50, mAP75). This strongly indicates that our method, by fully leveraging the information hidden in each spatialized frequency band map and interacting with spatial information via cross-domain attention fusion, allows RAW based object detection to be not only superior in dark environments but also robust in bright ones.

\begin{figure*}[h!]
    \centering
    \includegraphics[width=0.495\textwidth]{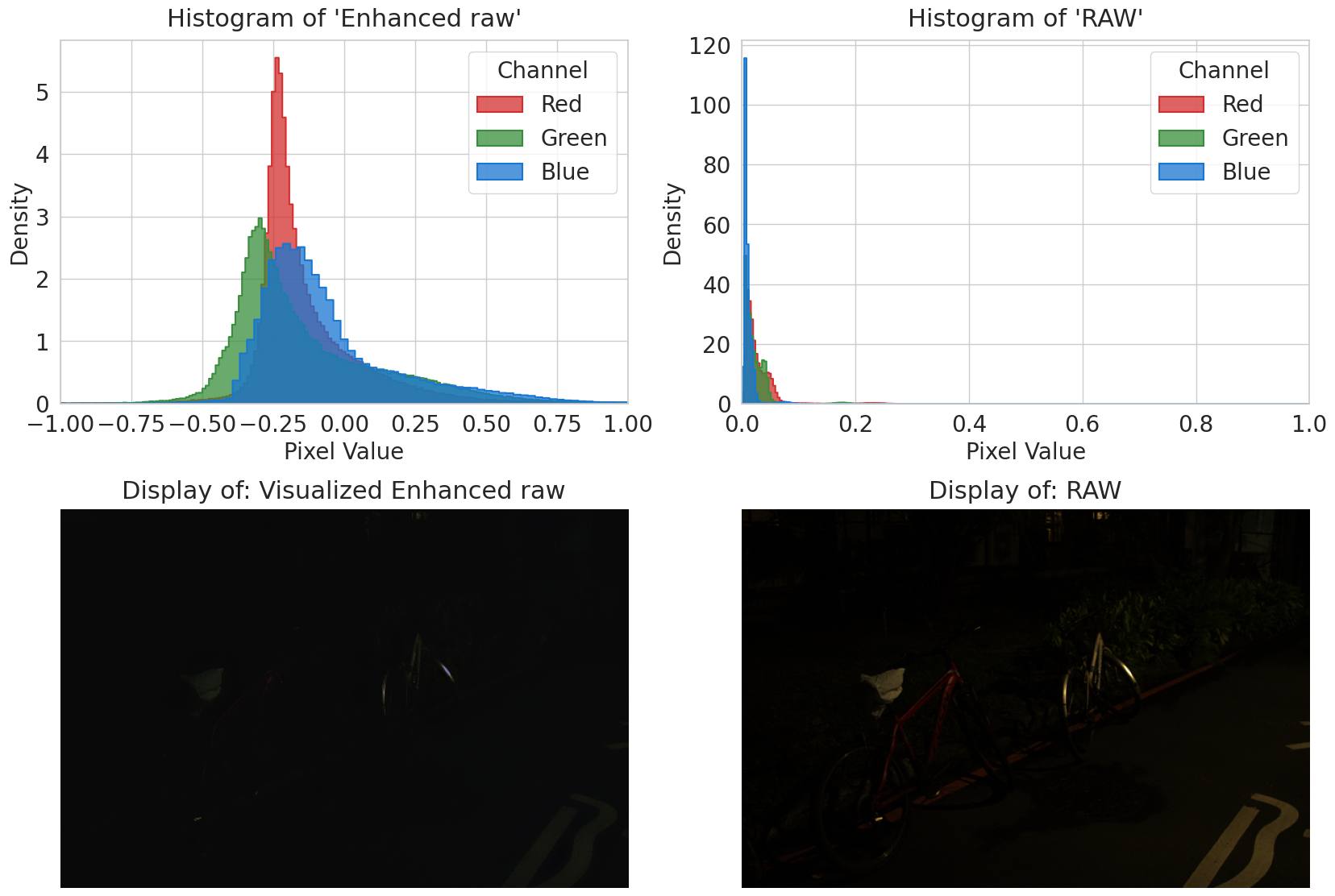}
    \includegraphics[width=0.495\textwidth]{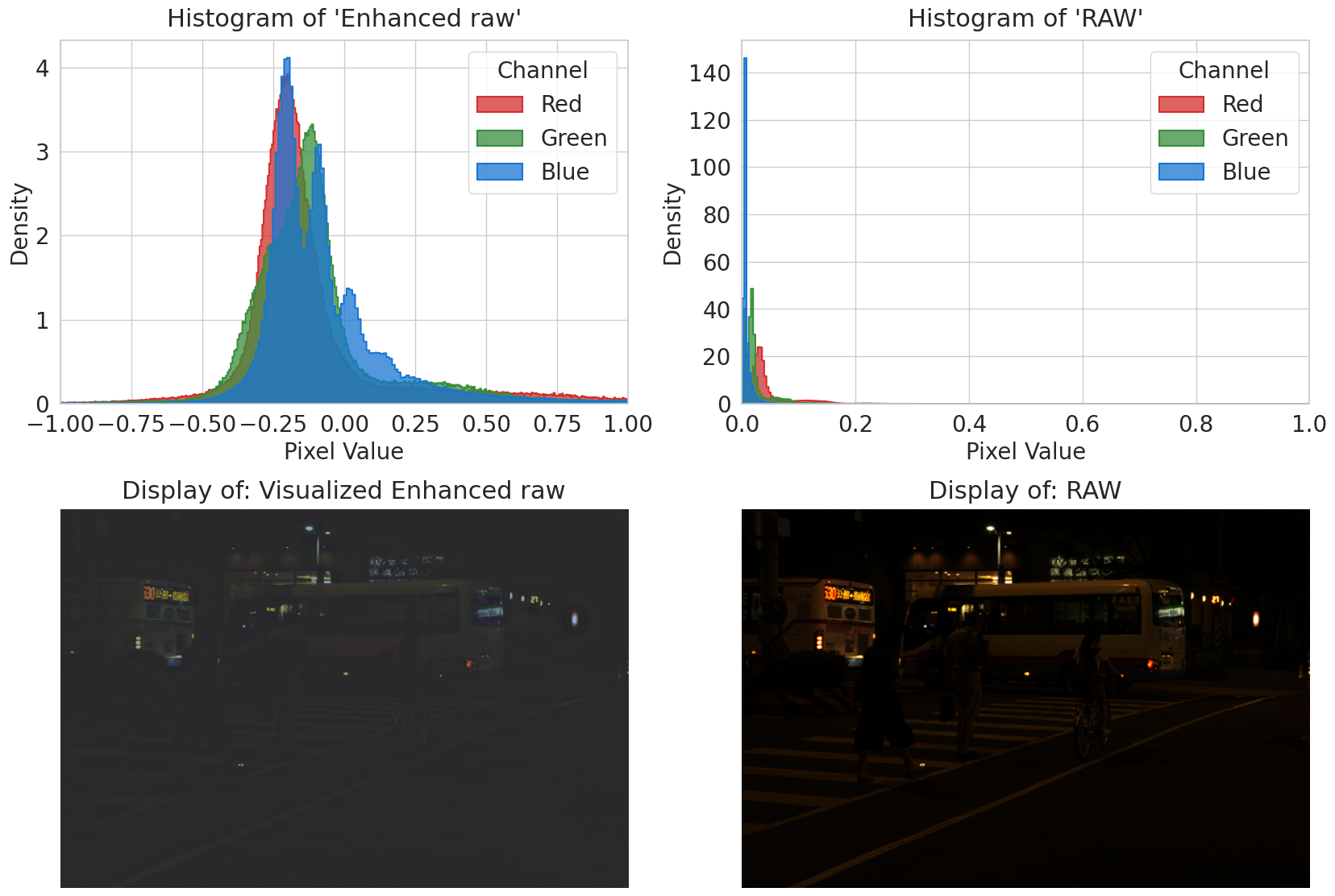}
    \caption{The Histogram Analysis of RAW and Enhanced RAW by our method. The shown images are from Nikon-Sony dataset. Red, Green and Blue curves are histogram of Red, Green, Blue channels respectively.}
    \label{fig:histograms}
\end{figure*}

As shown in Table \ref{tab:model_efficiency}, our leading precision is achieved with the second lowest parameters, which proves the efficiency of our model. This ``high precision and low parameter" features make it very suitable for resource limited environments.
Although the inference time of the current version is about 0.1s longer than that of the fastest method of its kind, we believe that this is a reasonable trade-off under the design of pursuing accuracy. The focus of our model is to verify the effectiveness of the architecture, and its low parameters provide great potential for future speed optimization.

\begin{table}[h] 
\centering
\small 
\setlength{\tabcolsep}{12pt} 
\begin{tabular}{@{}lccc@{}}
\toprule
\textbf{Method} & \textbf{Params (M)} & \textbf{Time (s)} & \textbf{mAP} \\
\midrule
Genisp          & 0.12 &  0.19 &  31.0 \\
Adaptive Module & 0.07 &  0.20 & 37.0 \\
Raw-Adapter   & 0.08 & 0.28 & 37.0 \\
IA-ISP & 0.03 & 0.25 & 38.0 \\
\midrule 
\textbf{Ours}   & 0.04 & 0.33 & 39.7 \\
\bottomrule
\end{tabular}
\caption{Comparison of size and inference time of different deep network based enhancement methods on LOD dataset.}
\label{tab:model_efficiency}
\end{table}

The histogram shown in Figure \ref{fig:histograms} indicates that, as described in \citep{li2024efficient} and \cite{raworcooked}, a critical reason for the inferior performance of using RAW image for object detection directly is that the pixel distribution of RAW image is highly concentrated. The histogram of RAW image here clearly shows the point. Moreover, this abnormal distribution buries the rich information of RAW image, resulting in the gradient vanishing effect, making it difficult for the model to learn and converge. The processed RAW image distributions produced by our method closely resemble a Gaussian distribution, indicating that our approach effectively harnesses the rich information inherent in RAW data and mitigates the gradient vanishing problem.

\section{Ablation Studies}
In order to verify the effectiveness of each core component in SF-CFAE and explore the influence of a different number of frequency bands, we conducted a series of ablation experiments. All experiments were performed on the LOD dataset using the RetinaNet detector with ResNet-50 backbone.
\begin{table}[h!]
\centering
\small
\setlength{\tabcolsep}{6pt} 
\begin{tabular}{@{}ccc|ccc@{}}
\toprule
\textbf{\shortstack{Spatial \\ Branch}} & \textbf{\shortstack{Frequency \\ Branch}} & \textbf{\shortstack{Cross- \\ Fusion}} & \textbf{mAP} & \textbf{mAP\textsubscript{50}} & \textbf{mAP\textsubscript{75}} \\
\midrule
\checkmark & \scalebox{0.85}[1]{$\times$} & \scalebox{0.85}[1]{$\times$} & 36.9 & 58.3 & 36.6 \\
 & \checkmark & \scalebox{0.85}[1]{$\times$} & 37.8 & \textbf{59.2} & 40.4 \\
\checkmark & \checkmark & \scalebox{0.85}[1]{$\times$} & 37.1 & \textbf{59.2} & 40.5 \\
\midrule
\checkmark & \checkmark & \checkmark & \textbf{39.7} & 59.1 & \textbf{44.9} \\
\bottomrule
\end{tabular}
\caption{We analyze the contributions of our key components on LOD dataset. \checkmark indicates the component is enabled.}
\label{tab:ablation_study_final}
\end{table}

In the component contribution analysis in Table \ref{tab:ablation_study_final}, we found that using the frequency domain branch (37.8 mAP) alone was better than that of using spatial branch (36.9 mAP), which proved that the spatialized frequency band map contains more effective information. However, without the cross-domain attention fusion module, the performance would be reduced to 37.1 mAP simply by using the features of the two branches. Finally, our complete model integrates all components and achieves the best performance (39.7 mAP), proving the effective synergy between various modules.
\begin{table}[h!]
\centering
\small
\setlength{\tabcolsep}{12pt} 
\begin{tabular}{@{}lccc@{}}
\toprule
\textbf{Number of Bands ($N$)} & \textbf{mAP} & \textbf{mAP\textsubscript{50}} & \textbf{mAP\textsubscript{75}} \\
\midrule
2 Bands & 37.8 & 58.1 & 39.1 \\
4 Bands & 38.2 & 58.6 & 40.9 \\
6 Bands & 38.8 & \textbf{59.6} & 40.3 \\
8 Bands (Ours) & \textbf{39.7} & 59.1 & \textbf{44.9} \\
10 Bands & 39.3 & 58.9 & 42.5 \\
\bottomrule
\end{tabular}
\caption{Ablation study on the number of frequency bands ($N$) on the LOD dataset.}
\label{tab:ablation_bands}
\end{table}

In the exploration of the number of bands $n$ in Table \ref{tab:ablation_bands}, we found that the performance reached its peak when $n = 8$. This indicates that a sufficient decomposition of the frequency domain is necessary, but excessive frequency bands (such as $n = 10$) leads to a slight drop in performance. Therefore, we chose $n=8$ as the standard configuration for all experiments.

\section{Conclusion}
In this paper, we study the application of RAW images in the object detection task. We propose a new method called Spatial-Frequency Aware enhancer (SFAE). The core of this method is that we decompose the image in the frequency domain and spatialize each frequency band for the network through iFFT and integrate the features in spatial and frequency domain by leveraging a cross-domain attention fusion module for multimodal synergy. Extensive experiments have shown that our method can stably improve the performance of a variety of detectors, surpassing existing SOTA methods. Ablation study further confirmed the necessity of the key components such as frequency domain branch and cross-domain attention fusion. However, we acknowledge that the computational complexity of our fusion module leads to a modest increase in inference time. We leave the optimization of this accuracy-speed trade-off as a direction for future work. Overall, our work provides a new and effective solution for RAW image processing oriented to machine vision.

\bibliography{aaai2026}

\end{document}